%% file: main.tex
\pdfoutput=1
\documentclass[11pt]{article}

\usepackage[preprint]{acl}

\usepackage{times}
\usepackage{latexsym}

\usepackage[T1]{fontenc}

\usepackage[utf8]{inputenc}

\usepackage{microtype}

\usepackage{inconsolata}

\usepackage{graphicx}

\usepackage{hyperref}       
\usepackage{url}            
\usepackage{booktabs}       
\usepackage{amsfonts}       
\usepackage{nicefrac}       
\usepackage{microtype}      
\usepackage{lipsum}
\usepackage{natbib}
\usepackage{algorithm}
\usepackage{algorithmicx}
\usepackage{algpseudocode}
\usepackage{amsmath}
\usepackage{fancyhdr}       
\usepackage{subfigure}
\usepackage{tabularx}
\usepackage{array}
\usepackage{xspace}
\usepackage{multirow}
\usepackage{multicol}
\usepackage{bm}
\usepackage{verbatim}
\usepackage[most]{tcolorbox}
\usepackage{mathtools}
\usepackage{fontawesome}

\input{math_commands}

\newcommand{\methodname}{{\sc FACT}\xspace}

\definecolor{prompt}{RGB}{223, 223, 192}
\definecolor{prompt-frame}{RGB}{137, 137, 90}
\definecolor{prompt2}{RGB}{223, 223, 192}
\definecolor{prompt2-frame}{RGB}{137, 137, 90}
\definecolor{prompt3}{RGB}{212, 238, 179}
\definecolor{prompt3-frame}{RGB}{117, 146, 77}
\definecolor{prompt4}{RGB}{212, 238, 179}
\definecolor{prompt4-frame}{RGB}{117, 146, 77}
\definecolor{prompt5}{RGB}{212, 238, 179}
\definecolor{prompt5-frame}{RGB}{117, 146, 77}
\definecolor{whitesmoke}{RGB}{245, 245, 245}  

\tcbset{
    prompt/.style={
        enhanced,
        breakable,
        colback=prompt,        
        colframe=prompt-frame,            
        boxrule=1pt,                       
        arc=4mm,                           
        left=1mm, right=1mm, top=1mm, bottom=1mm, 
        boxsep=4pt,                        
        before skip=10pt, after skip=10pt, 
        overlay={
            \node[anchor=north west, xshift=4pt, text=white] at (frame.north west) {\faDatabase};
        },
        title={\textbf{Prompt used for \methodname{}'s retrieval step for the QA tasks}},
        coltitle=white,
        fonttitle=\bfseries\small
    }
}
\tcbset{
    prompt_ruler/.style={
        enhanced,
        breakable,
        colback=prompt2,        
        colframe=prompt2-frame,            
        boxrule=1pt,                       
        arc=4mm,                           
        left=1mm, right=1mm, top=1mm, bottom=1mm, 
        boxsep=4pt,                        
        before skip=10pt, after skip=10pt, 
        title={\textbf{Prompt used for \methodname{}'s retrieval step for the RULER Retrieval tasks}},
        coltitle=white,
        fonttitle=\bfseries\small
    }
}

\tcbset{
    prompt_cs/.style={
        enhanced,
        breakable,
        colback=prompt2,        
        colframe=prompt2-frame,            
        boxrule=1pt,                       
        arc=4mm,                           
        left=1mm, right=1mm, top=1mm, bottom=1mm, 
        boxsep=4pt,                        
        before skip=10pt, after skip=10pt, 
        title={\textbf{Prompt used for \methodname{}'s retrieval step for the Counting-Stars tasks}},
        coltitle=white,
        fonttitle=\bfseries\small
    }
}

\tcbset{
    prompt_mqniah/.style={
        enhanced,
        breakable,
        colback=prompt2,        
        colframe=prompt2-frame,            
        boxrule=1pt,                       
        arc=4mm,                           
        left=1mm, right=1mm, top=1mm, bottom=1mm, 
        boxsep=4pt,                        
        before skip=10pt, after skip=10pt, 
        title={\textbf{Prompt used for MQ-NIAH task in Section~\ref{sec:challenge}}},
        coltitle=white,
        fonttitle=\bfseries\small
    }
}

\title{FACT: Examining the Effectiveness of Iterative Context Rewriting\\ for Multi-fact Retrieval}

\author{
 \textbf{Jinlin Wang\textsuperscript{1}$\footnotemark[1]$},
 \textbf{Suyuchen Wang\textsuperscript{2}$\thanks{Equal contribution.}$},
 \textbf{Ziwen Xia\textsuperscript{1}},
 \textbf{Sirui Hong\textsuperscript{1}},
 \textbf{Yun Zhu\textsuperscript{3}},\\
 \textbf{Bang Liu\textsuperscript{2}$^\S$},
 \textbf{Chenglin Wu\textsuperscript{1}$^\S$}
\\
 \textsuperscript{1}DeepWisdom\quad
 \textsuperscript{2}Universit\'e de Montr\'eal \& Mila\quad
 \textsuperscript{3}Google
\\
}

\begin{document}
\maketitle

\renewcommand{\thefootnote}{\S}
\footnotetext[2]{Correspondance: Bang Liu (\href{mailto:bang.liu@umontreal.ca}{bang.liu@umontreal.ca}) and Chenglin Wu (\href{mailto:alexanderwu@deepwisdom.ai}{alexanderwu@deepwisdom.ai}).}

\begin{abstract}

\input{files/1-abstract}
\end{abstract}

\input{files/2-introduction}

\input{files/4-methodology}

\input{files/5-experiment}

\input{files/6-conclusion}

\input{files/7-limitation}

\bibliography{custom,references_new}

\newpage
\appendix

\input{files/8-appendix}



\end{document}

%% file: math_commands.tex
\def\eqref#1{equation~\ref{#1}}

\def\1{\bm{1}}

\def\vx{{\bm{x}}}

\def\mC{{\bm{C}}}

\DeclareMathAlphabet{\mathsfit}{\encodingdefault}{\sfdefault}{m}{sl}
\SetMathAlphabet{\mathsfit}{bold}{\encodingdefault}{\sfdefault}{bx}{n}

\def\sV{{\mathbb{V}}}

\newcommand{\R}{\mathbb{R}}

\usepackage{cancel}

%% file: files/1-abstract.tex
Large Language Models (LLMs) are proficient at retrieving single facts from extended contexts, yet they struggle with tasks requiring the simultaneous retrieval of multiple facts, especially during generation. This paper identifies a novel ``lost-in-the-middle'' phenomenon, where LLMs progressively lose track of critical information throughout the generation process, resulting in incomplete or inaccurate retrieval. To address this challenge, we introduce Find All Crucial Texts (\methodname{}), an iterative retrieval method that refines context through successive rounds of rewriting. This approach enables models to capture essential facts incrementally, which are often overlooked in single-pass retrieval. Experiments demonstrate that \methodname{} substantially enhances multi-fact retrieval performance across various tasks, though improvements are less notable in general-purpose QA scenarios. Our findings shed light on the limitations of LLMs in multi-fact retrieval and underscore the need for more resilient long-context retrieval strategies.

%% file: files/2-introduction.tex
\section{Introduction}

Large Language Models (LLMs) have demonstrated impressive capabilities in various NLP tasks, particularly in situations where single, salient facts need to be retrieved from a long context~\citep{shi2023replug,izacard2021leveraging,jiang2023active,lin2023ra,jeong2024adaptive}. These ``needle-in-a-haystack'' ~\citep{gkamradt2023llmtest} tasks highlight the strength of modern LLMs in isolating critical information~\citep{hsieh2024ruler,yoran2023making}. However, in tasks requiring the retrieval of multiple facts simultaneously—referred to as \textit{multi-fact retrieval tasks}—the performance of both open-source and proprietary LLMs noticeably degrades~\citep{hsieh2024ruler,li2024needlebench}. This is particularly problematic in long-context scenarios~\citep{liu2023lost}, where models struggle to retain and retrieve multiple pieces of key information, leading to incomplete or erroneous results.

To improve LLMs' performance in multi-fact retrieval tasks, we conducted an analysis of the failure patterns specific to this context. Our findings reveal that the core issue is not identifying relevant information individually but the model's difficulty in focusing on multiple facts as they accumulate. Therefore, in multi-fact retrieval scenarios, as the generation process goes on, the model gradually loses track of the information to be retrieved and tends to retrieve incomplete or incorrect information. This issue, which we term the ``lost-in-the-middle'' in multi-fact retrieval generation, occurs when multiple critical pieces of information are distributed throughout the context. Conventional retrieval techniques—whether relying on LLM querying or vector-based methods—tend to focus on isolated facts, missing the broader context needed to retrieve all necessary information for complete understanding or reasoning.

Based on this observation, we investigate whether a multi-round retrieval scheme can mitigate performance drops in multi-fact retrieval tasks. Specifically, we introduce \textbf{Find All Crucial Texts (\methodname{})}, an iterative approach tailored for multi-fact retrieval. In our method, ``context rewriting'' leverages previously retrieved information to iteratively refine the context. Single-pass retrieval often fails to capture multiple facts, as the model’s attention tends to focus primarily on the top-ranked fact. Our iterative process addresses this limitation by progressively removing identified facts from the context, allowing the model to concentrate on additional critical facts in subsequent rounds.

We empirically demonstrate that~\methodname{} significantly outperforms baseline methods in retrieving multiple important facts from long contexts. However, we also show mixed results when applying \methodname{} on general-purpose QA tasks, where we analyze the influence of rewriting rounds, model families, model sizes, and task types on the performance of \methodname{}-like iterative rewriting methods. We conclude that although the multi-round retrieval method drastically benefits retrieval tasks, this performance boost is not universally transferrable to other tasks or models. Our research demonstrates that retrieval tasks themselves are not enough for evaluating long-context scaling methods and calls for better context-building mechanisms, long-context reasoning approaches, and more agentic long-context solutions.

%% file: files/4-methodology.tex
\section{The Challenge of Multi-fact Retrieval}
\label{sec:challenge}

In~\citet{hsieh2024ruler}, models demonstrate a consistent decline in performance as the number of required retrievals from context increases. This section aims to explore the underlying mechanisms behind this degradation: does the model prematurely terminate its retrieval process, or does it struggle to track and process the necessary information?

We approach this through a mechanistic analysis inspired by~\citet{lu2024insights}. Specifically, we adopt the multi-query needle-in-a-haystack (MQ-NIAH) task from RULER~\citep{hsieh2024ruler}, where the model is presented with a context of key-value pairs and a question containing multiple keys, tasked with retrieving the corresponding values sequentially. To diagnose the model’s internal representations, we train a linear probe on each layer of a LLaMA-3 8B Instruct model. The probe maps the intermediate layer representation, $\vx$, corresponding to an output position of a value, to an output value token $y$. The probe’s accuracy reflects the degree to which the model's internal state retains the necessary information to output the correct value, allowing us to distinguish between cases where the model ``knows'' the information but fails to output it (high probe accuracy) and cases where the model has entirely lost track of the required information (low probe accuracy). For the details of linear probe training, please refer to Appendix~\ref{appx:probe}.

\begin{figure}
    \centering
    \includegraphics[width=\linewidth]{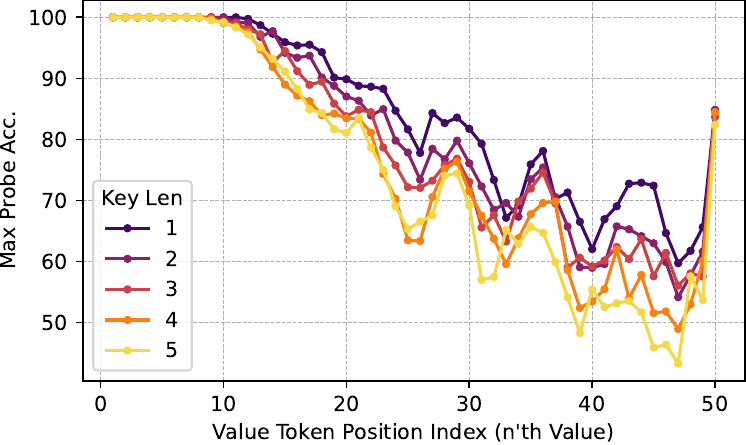}
    \caption{Maximum probing accuracy in a multi-query needle-in-a-haystack (MQ-NIAH) task across all layers of a Llama-3 8B Instruct model. The figure shows a ``lost-in-the-middle'' phenomenon for the \textit{generation} process in MQ-NIAH.}
    \label{fig:probing_acc}
    \vspace{-3mm}
\end{figure}

Figure~\ref{fig:probing_acc} plots the maximum probe accuracy against output position for different query lengths in a 50-query MQ-NIAH setting. The results reveal a clear \textbf{``lost-in-the-middle'' pattern \textit{during generation}}: as the generation progresses, the model progressively loses information until it recovers at the final few generated tokens.
Please note that this is different from the ``lost-in-the-middle'' \textit{in context}~\citep{liu2024lost}, where it focus on single information in the middle of the context instead.
Notably, the position of the accuracy turning point is largely invariant to the number of key tokens, suggesting that the performance degradation is not due to an overloaded number of key tokens. This pattern implies a fundamental constraint in the model's capacity: \textbf{it appears unable to reliably retrieve and track more than a certain number of key pieces of information concurrently from the context}.

\section{The~\methodname{} Method}
As noted above, the facts are basic constituent units within the context. They can be used to provide information in retrieval tasks and to generate answers in QA tasks. The completeness of facts is crucial to retrieval and QA performance. To this end, we introduce an iterative rewriting method called \methodname, which significantly enhances fact retrieval performance in common scenarios.
This largely addresses the challenges mentioned in Section~\ref{sec:challenge}.

\subsection{Iterative Rewriting}
To solve the problem of incomplete or inaccurate facts, we employ an iterative rewriting approach for fact retrieval. Specifically, based on the user's query, candidate facts are retrieved through methods such as using LLMs as retrievers or vector-based approaches. These candidate facts are then located inside the context, where they are rewrited by either removing or replaced with other noise data, resulting in a new context. This process is repeated until reaching the maximum number of iterations or meeting the stopping criteria. The candidate facts found in each iteration are aggregated to form the final set of facts. Algorithm \ref{algo:fact} describes the complete process in detail.

\begin{algorithm}
\caption{\methodname}
\small
\label{algo:fact}
\begin{algorithmic}[1]
\Require $Q$: the user query text, $C$: the context of the sample, $n$: number of iterations, $\mathtt{Retrieve}$: the retrieval function, $\mathtt{Rewrite}$: the context rewriting function, $\mathtt{Stop}$: the iteration stop judgment function
\Ensure $F$: the set of final found facts

\State $F = []$
\For{$i = 1$ to $n$}
    \State $cand\_facts$ = $\mathtt{Retrieve}$($Q$, $C$)
    \State $C$ = $\mathtt{Rewrite}$($cand\_facts$, $C$)
    \State $F$.extend($cand\_facts$)
    \If{$\mathtt{Stop}$($F$, $C$)}
        \State \textbf{break}
    \EndIf
\EndFor
\State \Return $F$

\end{algorithmic}
\end{algorithm}

%% file: files/5-experiment.tex
\section{Experiments}
\label{sec:experiments}

\subsection{Settings}

We test the performance of~\methodname~equipped with closed-sourced GPT-4o and GPT-4o-mini~\citep{gpt4o}\footnote{We adopt the \texttt{gpt-4o-2024-08-06} and \texttt{gpt-4o-mini-2024-07-18} versions in our experiments.}, and open-sourced Llama-3.1 8B Instruct~\citep{dubey2024llama3}. We report the performance on two types of tasks:
\begin{itemize}
    \item \textbf{Retrieval Tasks}, where the model directly retrieves multiple key information in the context. This includes RULER~\citep{hsieh2024ruler} and Counting Stars~\citep{song2024countingstars}.
\item \textbf{QA Tasks}, which requires reasoning about the provided context. This type of task includes: (1) Single-doc QA tasks, including NarrativeQA~\citep{kocisky2018narrative}, Qasper~\citep{dasigi2021dataset}, and MultiFieldQA~\citep{bai2024longbench}; (2) Multi-doc QA tasks, including HotpotQA~\citep{yang2018hotpotqa}, 2WikiMQA~\citep{ho2020constructing}, and MuSiQue~\citep{trivedi2022musique}. The QA tasks adopt the contexts and prompt templates from LongBench~\cite{bai2024longbench}. Please refer to Appendix~\ref{appx:statistics} for the statistics of the QA tasks.
\end{itemize}

In the experiments, we compare the results of \methodname{} against a baseline direct retrieval method for each model. This direct retrieval setting returns all the retrieved information or directly answers the question in one shot with the default prompt for each task. We include the prompts we used for the retrieval task and the retrieval step of the QA tasks in Appendix~\ref{appx:prompts}.

\subsection{Retrieval Tasks}

We present the results of the retrieval tasks in Table~\ref{tab:retrieval_results}. The results demonstrate a significant improvement in retrieval performance when applying our proposed method across both open-source and closed-source models. Across all tasks, the method consistently enhances the models' ability to retrieve relevant information from long contexts, outperforming the direct retrieval baselines substantially. This is particularly evident in tasks with longer context lengths, where traditional retrieval methods struggle. Moreover, for better-performing models like GPT-4o and GPT-4o-mini, \methodname{} achieves nearly perfect results.

\begin{figure}
    \centering
    \includegraphics[width=\linewidth]{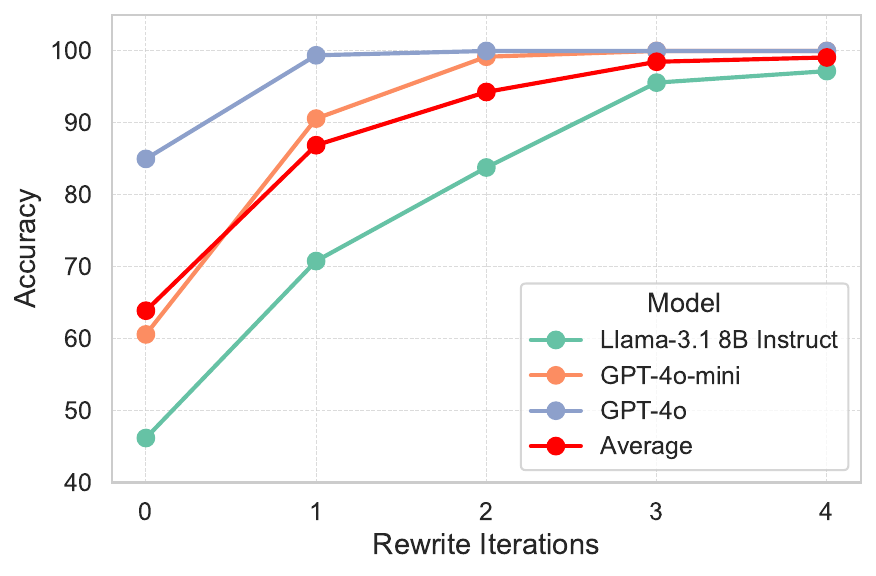}
    \caption{Retrieval Task performances under different numbers of rewriting iterations. The red line denotes the average performance across all tasks.}
    \label{fig:altering_retrieval_iters}
    \vspace{-1mm}
\end{figure}

\begin{table}[t]
\centering
\small

\resizebox{\columnwidth}{!}{
\begin{tabular}{ll|cc|cc|cc|c}
    \toprule
    & & \multicolumn{4}{c|}{\textbf{RULER}} & \multicolumn{2}{c|}{\textbf{Counting Stars}} & \multirow{3}{*}{\textbf{Overall}} \\
    \cmidrule(lr){3-6} \cmidrule(lr){7-8}
    \multirow{3}{*}{\textbf{LLM}} & \multirow{3}{*}{\textbf{Method}} & \multicolumn{2}{c|}{\textbf{K1V10Q1}} & \multicolumn{2}{c|}{\textbf{K5V10Q1}} & \multicolumn{2}{c|}{\textbf{N32}} &  \\
    \cmidrule(lr){3-8}
    & & 4K & 16K & 4K & 16K & 4K & 16K & \\
    \midrule
    \multirow{2}{*}{L. 8B}
        & base. & 70.0 & 46.2 & 80.2 & 61.0 & 98.6 & 80.9 & 72.8 \\
        & \textbf{\methodname{}} & \textbf{98.4} & \textbf{83.8} & \textbf{100.0} & \textbf{98.6} & \textbf{100.0} & \textbf{99.2} & \textbf{96.7} \\
    \midrule
    \multirow{2}{*}{4o-mini}
        & base. & 73.4 & 60.6 & 81.6 & 59.6 & 96.7 & 72.2 & 70.0 \\
        & \textbf{\methodname{}} & \textbf{99.8} & \textbf{99.2} & \textbf{100.0} & \textbf{99.6} & \textbf{98.0} & \textbf{99.9} & \textbf{99.4} \\
    \midrule
    \multirow{2}{*}{gpt-4o}
        & base. & 98.4 & 85.0 & 99.8 & 80.6 & 99.8 & 92.7 & 92.7 \\
        & \textbf{\methodname{}} & \textbf{100.0} & \textbf{100.0} & \textbf{100.0} & \textbf{99.6} & \textbf{100.0} & \textbf{100.0} & \textbf{99.9} \\
    \bottomrule
\end{tabular}
}
\caption{Performance Comparison on Retrieval Tasks. ``K$x$V$y$Q$z$'' denotes adding $x$ needles inside the context and retrieving $y$ values from a single query or $z$ queries. ``N$y$'' denotes retrieving $y$ needles from the context. The best performances of each model on each task are \textbf{bolded}. ``L. 8B'' denotes Llama-3.1 8B Instruct; ``4o-mini'' denotes GPT-4o-mini.}
\label{tab:retrieval_results}
\end{table}

In Figure~\ref{fig:altering_retrieval_iters}, we show the comparison of our proposed~\methodname{} against baselines under varied iterations with 16K context length in RULER K1V10Q1. Note that as the iteration increases, the overall scores steadily rise. This is especially true for Llama-3.1 8B Instruct: when the iteration reaches 3, the score increases by nearly 50 percentage points.

\begin{figure*}
    \centering
    \includegraphics[width=\linewidth]{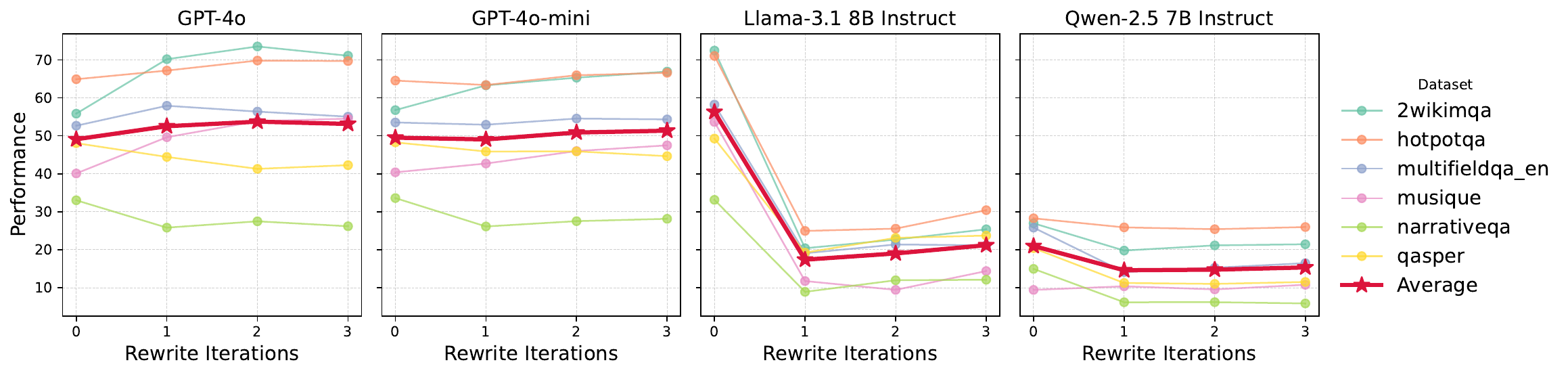}
    \caption{QA Task performances under different numbers of rewriting iterations. The red line denotes the average performance across all tasks.}
    \label{fig:qa}
    \vspace{-3mm}
\end{figure*}

\subsection{Question Answering Tasks}

We present the results of QA Tasks in Figure~\ref{fig:qa}.

\paragraph{Effect of iterative context rewriting on different model families.} We include Qwen-2.5 7B~\citep{yang2024qwen2} into discussion in this section. The performance impact of iterative context rewriting varies significantly across model families. GPT-4o and GPT-4o-mini consistently improve as the number of rewriting iterations increases. However, Llama-3.1 and Qwen-2.5 show a noticeable performance decline with iterative retrieval, particularly Llama-3.1, which struggles with retrieved context. This difference likely comes from training differences: GPT-4o may have been specifically trained on retrieval-augmented tasks, while Llama-3.1 and Qwen-2.5 may lack such training, making them more prone to hallucinations or errors.

\paragraph{Iterative rewriting versus one-shot retrieval.} Our results show that iterative rewriting outperforms one-shot retrieval, especially for models suited to retrieval-based tasks. Iterative rewriting leads to continuous improvements across iterations, showing the benefits of gradual context refinement. This supports our hypothesis in Section~\ref{sec:challenge}, which suggests that repeated enhancement of retrieved context improves model understanding and response quality.

\paragraph{Variability in task-specific performance with iterative rewriting.} The impact of iterative rewriting varies significantly across tasks. For the GPT-4o family, we see major gains for datasets like \textit{2wikimqa} and \textit{MuSiQue} but minor declines for \textit{Qasper} and \textit{NarrativeQA}. This is likely due to the different characteristics of each dataset. \textit{2wikimqa} and \textit{MuSiQue} contain dense factual information, which benefits from iterative rewriting by emphasizing key details and reducing noise, thereby improving accuracy. On the other hand, \textit{Qasper} and \textit{NarrativeQA} require nuanced reasoning and complex knowledge, which are beyond mere retrieval. Iterative rewriting in these cases may oversimplify or alter essential information, leading to loss of detail and increased ambiguity. Thus, while factual tasks benefit from \methodname{}, highly structured or narrative tasks may not.

%% file: files/6-conclusion.tex
\section{Conclusion}

This paper explored the challenges faced by Large Language Models (LLMs) in multi-fact retrieval tasks, particularly the ``lost-in-the-middle'' phenomenon, where models progressively lose track of key facts during generation. To address this, we introduced \methodname{}, an iterative context-rewriting method designed to improve multi-fact retrieval by progressively refining context. Our experiments show that \methodname{} significantly boosts retrieval performance in long-context scenarios, though results were mixed for general-purpose QA tasks.

These findings underscore the need for robust retrieval mechanisms that go beyond single-pass methods, highlighting the value of iterative refinement in complex retrieval settings. While \methodname{} proves effective for fact-intensive retrieval, its mixed performance on QA tasks suggests further research is needed to adapt iterative methods for broader NLP contexts. Future work should explore dynamic rewriting techniques tailored to task characteristics, balancing context enrichment with the retention of essential information. This could include dataset-aware rewriting strategies that adjust context modification based on task demands, optimizing performance while minimizing trade-offs. Additionally, task-specific training focused on retrieval could enhance the efficacy of iterative context rewriting.
Overall, this work lays a foundation for advancing context-building and long-context reasoning methods, pushing the boundaries of multi-fact retrieval capabilities in LLMs.

%% file: files/7-limitation.tex
\section*{Limitations}

This short paper includes the insights and findings of our experiments to improve LLMs' multi-fact retrieval performance. While the \methodname{} method shows considerable promise in improving multi-fact retrieval performance, there are several aspects that warrant further exploration, which we believe represent opportunities for future work rather than critical shortcomings.

\paragraph{Task-specific Performance Variability.}

\methodname{} exhibits significant improvements in multi-fact retrieval tasks, but its performance gains in general-purpose QA tasks are more mixed. This variation likely stems from the fundamental differences in task requirements: \methodname{} is particularly well-suited to fact-heavy retrieval tasks, where it refines the context over iterations. However, the iterative approach may not always lead to optimal outcomes in tasks requiring nuanced reasoning or comprehension, such as NarrativeQA or Qasper. Nonetheless, we see this as an opportunity to explore task-adaptive strategies that fine-tune the number of iterations or degree of context rewriting based on specific task characteristics.

\paragraph{Model-specific Behavior.}

The effectiveness of \methodname{} can differ across model families. Although closed-source models such as GPT-4o benefit significantly from iterative rewriting, some open-source models, such as LLaMA-3.1, show smaller gains or sometimes negative gains in retrieval tasks. This is likely due to differences in training regimes and architectures. However, these results highlight the potential to improve the performance of open-source models through targeted training in retrieval-augmented tasks. Addressing this presents an exciting avenue for future research, aiming to make \methodname{} more universally beneficial across various model types.

\paragraph{Computational Considerations.}

\methodname{} introduces additional computation due to its iterative nature, which could increase latency in certain applications. However, the trade-off between accuracy and computational overhead is a common challenge in advanced retrieval methods. In practice, this issue can be mitigated by fine-tuning the number of iterations or applying FACT selectively to tasks where its benefits justify the additional cost. Further research on optimizing the efficiency of iterative processes could help minimize this overhead.

\paragraph{Generalization to Broader NLP Tasks.}

\methodname{} is designed primarily for multi-fact retrieval tasks, and it excels in these tasks. Its application to more complex reasoning tasks, while promising, has room for improvement. We do not see this as a fundamental limitation of \methodname{}, but rather a natural constraint given its design focus. Adapting \methodname{} to tasks requiring deeper reasoning or synthesis remains an exciting challenge for future research, which could involve integrating more advanced reasoning or agentic procedures into the iterative process.

%% file: files/8-appendix.tex
\section{Dataset Statistics}
\label{appx:statistics}

\begin{table}[th]
\centering
\resizebox{\columnwidth}{!}{
\begin{tabular}{c|c|c|c|c|c|c}
\toprule
\textbf{Dataset} & \textbf{NQA} & \textbf{Qasper} & \textbf{MFQA} & \textbf{HotpotQA} & \textbf{2Wiki} & \textbf{MuSiQue}  \\
\midrule
\textbf{\#Samples} & 200 & 200 & 150 & 200 & 200 & 200 \\
\textbf{Avg Length} & 18,409 & 3,619 & 4,559 & 9,151 & 4,887 & 11,214  \\
\textbf{Metric} & F1 & F1 & F1 & F1 & F1 & F1 \\
\bottomrule
\end{tabular}
}
\caption{Statistics of the QA datasets.}
\label{tab:statistics}
\end{table}

In this section, we provide the dataset statistics for the QA datasets we used in Section~\ref{sec:experiments} in Table~\ref{tab:statistics}. These datasets are derived from LongBench~\citep{bai2024longbench}. For the retrieval datasets, please refer to the configurations specified in Table~\ref{tab:retrieval_results}.

\section{Prompts}
\label{appx:prompts}

For the retrieval tasks and QA tasks, we use the official prompt provied by RULER~\citep{hsieh2024ruler} or Counting Stars~\citep{song2024countingstars}, and LongBench~\citep{bai2024longbench}, respectively. We provide the prompt template we used for \methodname{}'s retrieval step for all the evaluated tasks below.

\begin{tcolorbox}[prompt_ruler]
    Some special magic numbers are hidden within the following text. Make sure to memorize it. I will quiz you about the numbers afterwards.\\

    \{context\}\\

    What are all the special magic numbers for \{query\} mentioned in the provided text? The special magic numbers for \{query\} mentioned in the provided text are
\end{tcolorbox}

\begin{tcolorbox}[prompt_cs]
    \{context1\}\\
    The little penguin counted \{number1\} \textasteriskcentered\\
    \{context2\}\\
    The little penguin counted \{number2\} \textasteriskcentered\\

    On this moonlit and misty night, the little penguin is looking up at the sky and concentrating on counting \textasteriskcentered. Please help the little penguin collect the number of \textasteriskcentered, for example: \{"little\_penguin": [x, x, x,...]\}. The summation is not required, and the numbers in [x, x, x,...] represent the counted number of \textasteriskcentered by the little penguin. Only output the results in JSON format without any explanation.
\end{tcolorbox}

\begin{tcolorbox}[prompt]
    Please retrieve all the sentences in the given documents that are important and relevant to answer the question.\\

    Question: \{question\} \\

    The following are given documents.\\

    \{context\}\\

    Please retrieve the sentences from the given documents that are relevant to answer the question. Do not repeat your generation. The question is highlighted again at below.\\

    Question: \{question\}\\

    Retrieved sentences:\\

    (For each retrieved sentence, please start from the bullet symbol "-", if no results, just return a single "-")
\end{tcolorbox}

\section{Linear Probe Training Details}
\label{appx:probe}

This section describes the training process for the linear probes used in Section~\ref{sec:challenge}, specifically for the MQ-NIAH task. This linear probe is a multi-fact retrieval extension of the one proposed by~\citet{lu2024insights}.

For the MQ-NIAH task introduced in Section~\ref{sec:challenge}, the model receives $n_q$ queries and must retrieve corresponding values from $n_k$ key-value pairs in the prompt, where each value consists of a single token. We define $\sV$ as the set of all possible single-token values. Given a prompt, we define the index of the token corresponding to the $i$-th output value as $t_i \in \R$, and the value token itself as $v_i \in \sV$.

Assume the LLM consists of $L$ layers. For each Transformer layer, we randomly initialize a linear classifier $\mC \in \R^{d \times v}$, where $d$ is the hidden dimension of the LLM, and $v = \left|\sV\right|$ is the number of possible values. Given the output from the $l$-th layer, denoted as $H_l \in \R^{L \times d}$ (with $L$ representing the sequence length), the linear classifier $\mC_l$ predicts the value $v_i$ using the hidden state $H_{l,[t_i,:]}$ for each $i \in \{1, \cdots, n_q\}$.

We collect training data and conduct inference using a specifically designed prompt. During training, we concatenate the ground-truth values to the prompt and record $v_i$ and $H_{l,[t_i,:]}$ for all layers $l \in \{1, \cdots, L\}$ and queries $i \in \{1, \cdots, n_q\}$ in a single forward pass. The training prompt is structured as follows:

\begin{tcolorbox}[prompt_mqniah]
    Extract the value corresponding to the specified key in the JSON object below.\\

    \{``|''\_separated\_keys\}\\

    JSON data: \{json\_formatted\_key\_value\_pairs\}\\

    Keys: \{``|''\_separated\_keys\}\\

    Corresponding Value:\\
\end{tcolorbox}

In our experiments, we use Llama-3 8B Instruct~\citep{dubey2024llama3} as the LLM, and we assign $n_q=50$, $n_k=200$. The linear classifiers are trained with the hyperparameters specified in Table~\ref{tab:linear_probe_training_detail}. All experiments are done with either a single NVIDIA RTX3090 24G or a single NVIDIA A100 40G.

\begin{table}[th]
\centering
\resizebox{\columnwidth}{!}{
\begin{tabular}{c|c|c|c}
\toprule
\textbf{Key}   & \#Samples & Epoch & Learning Rate   \\
\midrule
\textbf{Value} & 2000      &  150  & 0.005  \\
\bottomrule
\end{tabular}
}
\caption{Hyperparameters of Linear Probe Training.}
\label{tab:linear_probe_training_detail}
\end{table}